\documentclass[letterpaper, 10 pt, conference]{ieeeconf}  
\usepackage{tikz}
\usepackage{colortbl}
\usepackage[normalem]{ulem}
\usetikzlibrary{positioning,fit,calc,backgrounds}
\usetikzlibrary{shapes,arrows}
\usepackage{amsmath,bm,times}

\usepackage{circuitikz}  

\usepackage{mathtools}

\usepackage{tikz,lipsum,lmodern}

\usepackage{graphicx}
\usepackage{caption}
\usepackage{booktabs}
\usepackage{subfig}
\usepackage{cite}

\usepackage{url}

\usetikzlibrary{graphs}
\usepackage{textcomp}
\usepackage{filecontents}
     
\usepackage{SIunits}      
                                           
\makeatletter
\newcommand*{\rom}[1]{\expandafter\@slowromancap\romannumeral #1@}
\makeatother
\IEEEoverridecommandlockouts                      
\title{\LARGE \bf 
Understanding of Object Manipulation Actions Using Human Multi-Modal Sensory Data}

\author{Bahareh Abbasi, Ehsan Noohi, Sina Parastegari, and Milo\v s \v Zefran}

\begin{document}
\maketitle
\thispagestyle{empty}
\pagestyle{empty}

\begin{abstract}

Object manipulation actions represent an important share of the Activities of Daily Living (ADLs). In this work, we study how to enable service robots to use human multi-modal data to understand object manipulation actions, and how they can recognize such actions when humans perform them during human-robot collaboration tasks. The multi-modal data in this study consists of videos, hand motion data, applied forces as represented by the pressure patterns on the hand, and measurements of the bending of the fingers, collected as human subjects performed manipulation actions. We investigate two different approaches. In the first one, we show that multi-modal signal (motion, finger bending and hand pressure) generated by the action can be decomposed into a set of primitives that can be seen as its building blocks. These primitives are used to define 24 multi-modal primitive features. The primitive features can in turn be used as an abstract representation of the multi-modal signal and employed for action recognition. In the latter approach, the visual features are extracted from the data using a pre-trained image classification deep convolutional neural network. The visual features are subsequently used to train the classifier. We also investigate whether adding data from other modalities produces a statistically significant improvement in the classifier performance. We show that both approaches produce a comparable performance. This implies that image-based methods can successfully recognize human actions during human-robot collaboration. On the other hand, in order to provide training data for the robot so it can learn how to perform object manipulation actions, multi-modal data provides a better alternative.
\end{abstract}
\section{introduction}
\label{Sec:intro}

Understanding human actions is a critical component in an effective human-robot interaction. To ensure successful collaboration, the robot should be able to recognize human actions, and produce appropriate physical behaviors. Service robots thus require both an action recognition module to identify the human actions and a learning module to learn how to replicate them. Development of such modules requires understanding how humans perform an action. To do so, it is necessary to collect data from various sensor modalities and understand their role. Among different actions of interest, object manipulation actions play key role in the ADLs. They are particularly challenging to study because they are difficult to measure in naturalistic settings. In this paper, we focus on modeling the human manipulation actions with the goal of enhancing the understanding between the human and the robot\par

In robotics, human manipulation actions have been traditionally studied either by using the video recordings~\cite{kjellstrom2008simultaneous,kjellstrom2011visual}, or by analyzing of human movement data~\cite{patel2013language,vicente2007action}. The extracted features can be exploited for modeling of human actions that involve interaction with objects, or for action recognition. The idea of breaking down these actions into action units analogous to phonemes in language, and to study the temporal sequence of these action units, has attracted considerable interest~\cite{vicente2007action,patel2013language,Sanmohan2009,kruger2010learning,khaleghi2015ic}. \par

The concept of action primitives can be used at a symbolic level to define the building blocks that can describe the actions. In turn, they can be used to train a low level model, like a support vector machine (SVM), to extract these primitives from the continuous motion trajectory, and a higher level temporal model such as Hidden Markov Model (HMM) to classify a sequence of them into actions~\cite{vicente2007action}. Alternatively, these two steps can be combined into a single hierarchical probabilistic temporal model (HHMM)~\cite{patel2013language}. In this approach, the association between the movement signal and the primitives is probabilistic due to variability seen among the subjects and even  within a subject~\cite{vicente2007action}, and human annotation of primitives/actions is required.
Another way of tackling this problem is to combine the movements at the trajectory level, and define the primitives based on repeated observed patterns~\cite{Sanmohan2009}. One of the drawback of this approach is that the obtained primitives are dependent on locations and other experimental setup properties. In \cite{kruger2010learning}, the problem is resolved by proposing a Parametric Hidden Markov Model (PHMM) for each primitive based on a quantization of object configuration space.\par

In this work, we model the human manipulation actions using two approaches, referred to as primitive-based and  visual-data-flow approach. We introduce the manipulation primitives which are inferred from the human multi-modal signals during the execution of these actions. The distinct sequential nature in which primitives occur in each manipulation action is exploited to recognize the action. The recognition module is thus based on a sequential model which can process a sequence of the primitives as they occur in time. A novel feature of our work is that we use force (more precisely, data from pressure sensors on the palm of the hand) as an additional information channel, i.e. the data includes force in addition to the hand velocity and bending of the finger joints. We define a set of 24 primitive features that can be easily identified in the multi-modal data, no learning or annotation needs to be used. These features are parameter-free and can be thus used as a symbolic representation of the multi-modal signal. Once the signals are converted into such a sequence of primitive features, the sequence can be used as an input for a sequential classifier for action recognition. Because these primitive features are parameter-free, the recognition is independent from the location of the action and generalizes across subjects.\par

The second approach uses visual features extracted from the raw data. We extract the raw visual features from recorded video using pre-trained deep convolutional neural network (DCNN). Training DCNN models from scratch requires a massive amount of data. Instead, we rely on the existing DCNNs that have been trained using ImageNet~\cite{russakovsky2015imagenet} as they are well suited for extracting the visual features. Subsequently, the vision-based features are utilized for training the classifier. 
We also train the same classifiers with the combination of the visual features and the multi-modal data to check whether adding the extra information significantly improves the recognition. To the best of our knowledge, we are among the first to use multi-modal data, in particular visual features along with motion and force data, for recognition of object manipulation actions. Finally, we show that vision can be successfully used for recognition and is statistically equivalent to multi-modal classifiers.\par


The rest of the paper is organized as follows, in section~\ref{Sect:related_works}, we review the literature on modeling human actions. In section.~\ref{Sect:actiontype}, we characterize the manipulative activities that are of interest to this work. Section \ref{Sect:primitves-model} is devoted to primitive-based approach. In section \ref{Sect:data-driven}, we describe our method for visual feature extraction and how to use them for recognition along with the non-vision features. The conclusion and future work is provided in section~\ref{sect:conclusion}.
\par

\begin{figure}[t]
\centering
\scalebox{0.8}{
\begin{tikzpicture}
\node[draw,minimum width=1cm, minimum height=1cm, align = center] (b) at (4,0) {\emph{\textbf{NatEXP}}  \\ Identify manipulation primitives};
\node[draw, minimum width=1cm, minimum height=1cm, align = center] (d) at (4,-1.5) {\emph{\textbf{MotionEXP}} \\Extract motion trajectories };
\node[draw, minimum width=1cm, minimum height=1cm, align = center] (f) at (4,-3) {\emph{\textbf{ValEXP}} \\ Performance validation};
\draw [-latex, thick] (b) --(d) node []{};
\draw [-latex, thick] (d) --(f) node []{};
\end{tikzpicture}
}
\caption{The three different human studies and their objectives in this work.}
\vspace{-6mm}
\label{fig:expdirectories}
\end{figure}
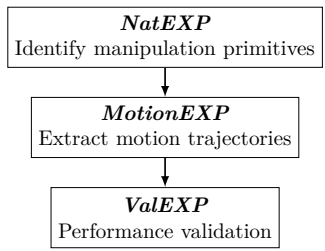

\section{related work}
\label{Sect:related_works}
There is a huge body of research on understanding human
activities. Many researchers have taken observational approach in modeling human activities and
proposed taxonomies for human actions~\cite{fleishman1984TaxonomiesHumanAbility}.
Pavlovic et al. \cite{pavlovic1997TaxonomiesManipulation} proposed a 
detailed taxonomy for hand and arm movements. They refer to all intentional
motions as ``gestures'' and suggest that human gestures are either
manipulative or communicative (e.g. mimetic or referential). Bullock
et al. \cite{bullock2013ManipulationTaxonomy} proposed a taxonomy with
15 classes for the manipulation tasks. Their taxonomy is based on the
type of the contact (prehensile or not) and the states of the hand
motion (no-motion, within-hand motion and non-within-hand
motion).

In contrast to the observational approach in modeling human actions,
many researchers have focused on proposing algorithms to determine
human actions from the measured physical signals, such as the motion
trajectories and applied forces/torques. Almost all of these works,
assume that a hierarchical relationship exists between the physical
signals and the high-level actions. One of the most appreciated models
for this hierarchical relation is the ``movement-activity-action'' model,
proposed by Bobick \cite{bobick1997MovementActivityAction}. He defines
movements as the most atomic primitives. Activity refers to a sequence
of such movements or states. 

While Bobick's model \cite{bobick1997MovementActivityAction} is
presented in the context of computer vision and image processing, the
idea of using primitive motions for determining human actions is
employed by many researchers in different research areas. For instance, 
Hogan and Sternad
\cite{hogan2013LimbPrimitives} suggested that human upper- and
lower-limb behaviors can be described by three classes of
``dynamic'' primitives: sub-movements (trajectory attractor),
oscillations (limit-cycle attractor) and mechanical impedances
(generalized stiffness).

\bgroup
\def\arraystretch{1}
\begin{table*}[t]
\centering
\begin{tabular}{|p{0.1\linewidth}|p{0.14\linewidth}|p{0.5\linewidth}|p{0.065\linewidth}|p{0.08\linewidth}|}
\hline
Action & Object & Activities & Instances & Duration (s) 
\\ \hline
\rowcolor[gray]{0.9} Putting on the shoe & \vspace{-2mm} shoe, subject's foot & Reaching to the object, grasping the object, picking up the object, reorienting the object (aligning with the foot), inserting the foot into the shoe & \vspace{-2mm} \hspace{4mm} 1 & \vspace{-2mm} \hspace{3mm} 32
\\ \hline

\rowcolor[gray]{0.9}Tying shoelace & \vspace{-2mm} shoelaces & Reaching the object, grabbing the object, performing the tying maneuvers, releasing the object  & \vspace{-2mm} \hspace{4mm} 1 & \vspace{-2mm} \hspace{3mm} 13 
\\ \hline

\rowcolor[gray]{0.9}Getting up & another person & Support other subject while getting up from a chair or a bed  & \hspace{4mm} 1 & \hspace{3.5mm} 6
\\ \hline

\rowcolor[gray]{0.9}Ambulating & another person & Support other subject while walking & \hspace{4mm} 1 & \hspace{3mm} 26 
\\ \hline

\vspace{-2mm} Open/Close &  door, closet, cabinet, fridge, drawer, ... & Reaching to the handle, grasping the handle, pushing/pulling the handle in the opening/closing direction, releasing the handle and reaching back & \vspace{-2mm} \hspace{3mm} 23 & \vspace{-2mm} \hspace{3mm} 57 
\\ \hline

Pick up &  plate, mug, spoon, ... & Reaching to the object, grasping the object and picking up the object  & \hspace{3mm} 24 & \hspace{3mm} 56 
\\ \hline

Place &  plate, mug, spoon, ... & Reaching to the goal, releasing the object and reaching back  & \hspace{3mm} 19 & \hspace{3mm} 43
\\ \hline

Carry/Hold & pot, plate, ... & Moving with/holding the grasped object & \hspace{3mm} 21 & \hspace{3mm} 70
\\ \hline

\rowcolor[gray]{0.9}Handover & plate, spoon, ... & Giving/receiving the object to the other person & \hspace{3mm} 18 & \hspace{3mm} 23
\\ \hline

Showing & bowl, plate, ... & Re-configuring the object to provide better view point for the other person & \hspace{4mm} 6 & \hspace{3mm} 12
\\ \hline

\vspace{-2mm} Displacing & plate, bowl, mug, silverware ...  & Reaching the object, grasping/touching the object, Sliding the object to target point, releasing the object, reaching back & \vspace{-2mm} \hspace{4mm} 7 & \vspace{-2mm} \hspace{3mm} 15
\\ \hline

\end{tabular}
\vspace{2mm}
\caption{The list of physical actions observed during the experiment. The  highlighted actions include non-habitual movements.}
\label{table:HumanStudyActions}
\vspace{-4mm}
\end{table*}
\egroup

Non-observational approaches for definition and utilization of the
primitives can be grouped into two main categories. The first category
consists of methods that introduce explicit primitives, usually in the
task space (or phase space). Therefore, these primitives usually have
physical representations and meanings. On the other hand, the methods
in the second category propose a set of data driven primitives,
usually in the feature space. These methods usually perform a
dimension reduction technique followed by applying a clustering
algorithm. Therefore, the primitives in this category usually do not
represent any physical entity.\par

In the first category, the primitive motions are either explicitly
proposed by the researchers or assumed to be available as a library
(e.g. pre-trained from human demonstrations). For instance,
Based on the idea
that human arm tends to move on a fixed plane, Fang and Ding
\cite{fang2013ExplicitPM} suggested explicit movements primitives
for anthropomorphic arms: ``moving on a working plane'', ``switching
different working planes'', ``translating the hand'', and ``rotating
the hand about a fixed axis''. 
Meier et al. \cite{meier2011DMP_PM_Segmentation} presented a
movement segmentation framework, assuming that a library of movement.\par

In the second category, the primitives are obtained in a machine
learning framework, without any prior knowledge or assumption. For
instance, to obtain primitive motions in a 3 to 5-second long
sequences of arm movements, Fod et al. \cite{fod2002PCAClustering}
used FastTrak motion tracking system. They segmented the velocity
signals using zero-crossing and thresholding techniques, performed PCA
on the segmented signals and applied K-means clustering on the first
30 principal components. Similarly, Lim et al. \cite{lim2005PCA_MP}
applied PCA on human motion capture data to obtain the joint
trajectory basis functions (movement primitives) and suggested that
arbitrary movements can be represented as a linear combination of these
basis functions. Jenkins et al. \cite{jenkins2002Isomap} used
spatio-temporal Isomap to identify clusters (primitive behaviors) in
human motion data. Williams et al. \cite{williams2006HandwritingPM}
employed factorial hidden Markov model to infer primitives in
handwriting data. Similarly, Inamura et
al. \cite{inamura2004MimesisTheory} employed expectation maximization
technique to learn the parameters of a Hidden Markov Model for unknown
primitive motions. Alvarez et al. \cite{alvarez2010LatentForce} used
second order linear differential equations in conjunction with a
latent force model to extract a sequence of dynamical motor
primitives. Our approach in defining the primitive falls in the first category. That is, by observing human activities of daily living (ADLs) in an experimental setup, we collected a list of building blocks that comprises the large proportion of the human manipulation actions. While most of the mentioned previous works processed the raw measured motion data for action recognition, our developed primitive-based recognition model rely on symbolic parameter free multi-modal primitive features and is generalized.

\section{human manipulation actions types}
\label{Sect:actiontype}

Depending on how the human nervous system handles the motion, arm/hand
movements can be categorized in three types: reflexive, habitual
and deliberate movements. Reflexive movements are motions whose control loops are implemented in the spinal cord and therefore, their response
to stimuli is very fast. An example of these
movements is the reflex of the arm when hand touches a hot object. In contrast, the habitual movements are
controlled at basal ganglia and are done as ordinary actions without any deliberation. On the other hand, certain manipulation tasks,
like spinning a basketball on a finger, are considered dexterous manipulation which need precise planning
in controller level. These manipulation tasks
are managed at prefrontal cortex and are referred to as deliberate
movements. Any new (unfamiliar) task is first treated as a deliberate
movement, but when it is repeated sufficiently often, it becomes
habitual.\par

Different people take different approaches in learning and performing
deliberate movements. As a result, the variability of the motion
trajectories between subjects for the same task is large. In contrast,
the motion profiles of the habitual movements are mainly governed by
the environment and human-body kinematics. Thus, the variability of the
motion trajectories between subjects is much smaller for
habitual movements. In this work, we focus on tasks that mainly
consist of habitual movements and exclude reflexive and deliberate
movements. ADLs are particularly good examples
of habitual movements and we will specifically study them in this
work.

\section{primitive-based approach}
\label{Sect:primitves-model}

In this section we describe our primitive-based approach. We will show that the habitual manipulation actions can be decomposed into a sequence of primitives and their corresponding features. Our work relies on three separate user studies (Fig.~\ref{fig:expdirectories}). In the \emph{NatEXP} study, data was collected in a fully functional apartment while an elder and a human helper performed ADLs. The \emph{MotionEXP} study was used to analyze motion trajectories of a subset of manipulation primitives to identify their primitive features. Finally, \emph{ValEXP} study was used to train the sequential models to validate the approach.\par
\subsection{NatEXP Human Study: Identify the Manipulation Primitives}
\label{Sect:natexp}

In our primary human study, \emph{NatEXP} (see Fig. \ref{fig:expdirectories}), we investigated the interaction between human and human, and collected data of interactions
between an elder and a human helper. The collected data was annotated for visual cues, spoken language, and manipulation actions. Here, we only focus on the manipulation actions and present that part of the corpus. Subsequently, we show that the substantial portion of these actions can be represented as the integration of proposed primitives in this section. The remaining parts of the corpus can be obtained from~\cite{chen2015RolesofHOAct}. 

The human study was conducted in a fully functional studio apartment at Rush University in Chicago. 
It consists of 15 sets of interactions in which gerontological nursing students provided care to the elder person. All elderly subjects were highly functioning at a cognitive level and did
not have any major physical impairment. 

Multi-modal data was collected during all
experiments. The room was equipped with seven cameras to ensure
multiple points of view. Both participants were wearing a microphone
to record their conversations, and a data glove on their dominant hand to collect haptics data.
For more details about the experimental setup see \cite{chen2015RolesofHOAct}. 
The videos are annotated and aligned with other signals using Anvil
\cite{kipp2001Anvil}. In total, there are 436 minutes of recorded signals, of which 266 minutes were the
helper-elder interactions during the ADLs. The rest was either
instructing the elder subject or setting up the experiment
equipment. \par

\begin{table}[t]
\centering
\scalebox{1}{
\begin{tabular}{l|c|c||c|c|}
\cline{2-5}
 & \# & \% & D & \%  \\ \hline
\multicolumn{1}{|l|} {Reaching movement} & 220 & 32.5\% & 162.32 & 62\%   \\ \hline
\rowcolor[gray]{.90}\multicolumn{1}{|l|} {Fixed-axis rotation} & 189 & 28\% & 94.44 & 36\% \\ \hline
\multicolumn{1}{|l|} {Grasp/release} & 130 & 19\% & NA & NA  \\ \hline
\rowcolor[gray]{.90}\multicolumn{1}{|l|} {Bend/extend} &131  &19.5 \ & NA & NA \\ \hline
\multicolumn{1}{|l|} {other} & 7 & 1\% & 3.8 & 2\%  \\ \hline
\end{tabular}
}
\caption{The proposed manipulation primitive movements and instances and duration (s) of each. }
\label{table:HumanStudyPrim}
\vspace{-4mm}
\end{table}

Each experiment involved interaction between an elder person and a
helper while performing four tasks: putting on shoes, getting up from
a bed or a chair, ambulating and preparing dinner. The ``putting on
shoes'' task consisted of the helper picking up the elder's shoe,
putting it on his/her foot and tying the laces. The ``getting up''
activity was performed by the helper supporting the elder's weight
while getting up from a chair or a bed. During ``ambulating'' action,
the helper provided support for the elder and helped the elder balance
while walking around the room. Finally, ``preparing dinner'' comprised
many sub-tasks, such as finding pots, opening cans and containers,
putting pots on a stove, setting up
the dinner table, and cleaning the table afterwards.

\bgroup
\def\arraystretch{1}
\begin{table}[t]
\centering
\scalebox{1}{
\begin{tabular}[H]{|c|}
\hline
\textbf{Manipulation Primitives}\\ \hline \hline
Reaching movement : \textit{reach}\\ 
\rowcolor[gray]{.90}Hand rotation:  \textit{rotate}\\ 
Grasp/Release: \textit{grasp/release}\\ 
\rowcolor[gray]{.90}Bend/Extend: \textit{bend/extend} \\ \hline

\end{tabular}}
\vspace{0mm}
\caption{The proposed manipulation primitives for representing the human manipulation actions.}
\label{table:primlist}
\vspace{-4mm}
\end{table}
\egroup
We analyzed in detail one of the experiments, focusing on manipulation actions. Of 19 minutes of recordings in that experiment, manipulation actions take 353 seconds. Table \ref{table:HumanStudyActions} shows a list of all the
manipulation actions that appear in the experiment. While most of the actions
comprised habitual movements, deliberate movements appeared in few
instances. For example, ``getting up'' and ``ambulating'' actions
require careful planning, because they involve ensuring the elder's
balance. Note that putting on shoes was also usually performed as a
sequence of planned deliberate movements because the subjects' feet
imposed unknown constraints for the motion planning. Similarly, tying
the shoelace on someone else's shoe was not habitual
movements. Table \ref{table:HumanStudyActions} describes different
activities (3rd column) that occur during an action (1st column). The last two columns of Table \ref{table:HumanStudyActions} give the
duration and frequency of the actions. In total, there were 122 actions. As illustrated, most of the actions comprised the habitual
movements.\par

Next, we attempted to identify primitives that could describe the manipulation actions. 
We observed that 81.35\% of the manipulation actions
can be described by only four activities: reaching
movement, 
hand rotation, bending/extending the fingers and
grasping/releasing the object. Table \ref{table:HumanStudyPrim} shows
the frequency and duration (in sec) of these primitives in the
experiment. Note that the first two primitives, reaching movement and hand rotation, can be
constrained by the environment (object). The manipulation actions that can not be described with the primitives above are mostly instances of deliberate movements. Curved hand trajectory due to an
obstacle and adjusting the pot's position on the stove are examples of
such movements. Based on these observations, we propose the human manipulation primitives that is listed in Table \ref{table:primlist}.

While the proposed primitives capture the experimental data, on their own they are not sufficient for recognizing actions in Table \ref{table:HumanStudyActions}; the primitives represent a level of abstraction that is too high and a sequence of primitives is not unique to a single action. This motivated us to use the primitives to identify features in the data that recognize different actions. Not only that, through the primitives, these data features can be defined explicitly as specific shapes in the signals.

\subsection{MotionEXP Human Study: Velocity Profile of Primitives}
In the previous section, we identified four manipulation
primitives in human manipulation actions, using the video
of the experiment. But in order to recognize different actions, we need to associate features to each primitive, and find a way to identify these features in the data. In this section, we describe how
a characteristic motion profile can be obtained for~\textit{reach} and~\textit{rotate} primitives. \par

Human motions have been studied extensively and it has been shown
that, for a ballistic motion, the speed profile is bell-shaped. Minimum-jerk model \cite{MinJerk_FlashHogans1985},
minimum torque change model \cite{MinTorqueChange_89,MinTorqueChange_99}, and minimum variance model
\cite{MinVar_Wolpert} are three popular models for this profile. Although the aforementioned models approximate the human reaching motion, we are interested in finding the motion trajectory not only for~\textit{reach} primitive but also for~\textit{rotate} primitive. We chose an empirical method to model these primitives to capture inter and intra subjects variation in performing these primitives with trying different conditions like empty hand, or holding different objects. 

\begin{figure}[t]
\vspace{-4mm}
  \subfloat[\textit{
reach} primitive \label{subfig-1:V}]{%
	\includegraphics[width=.47\columnwidth]{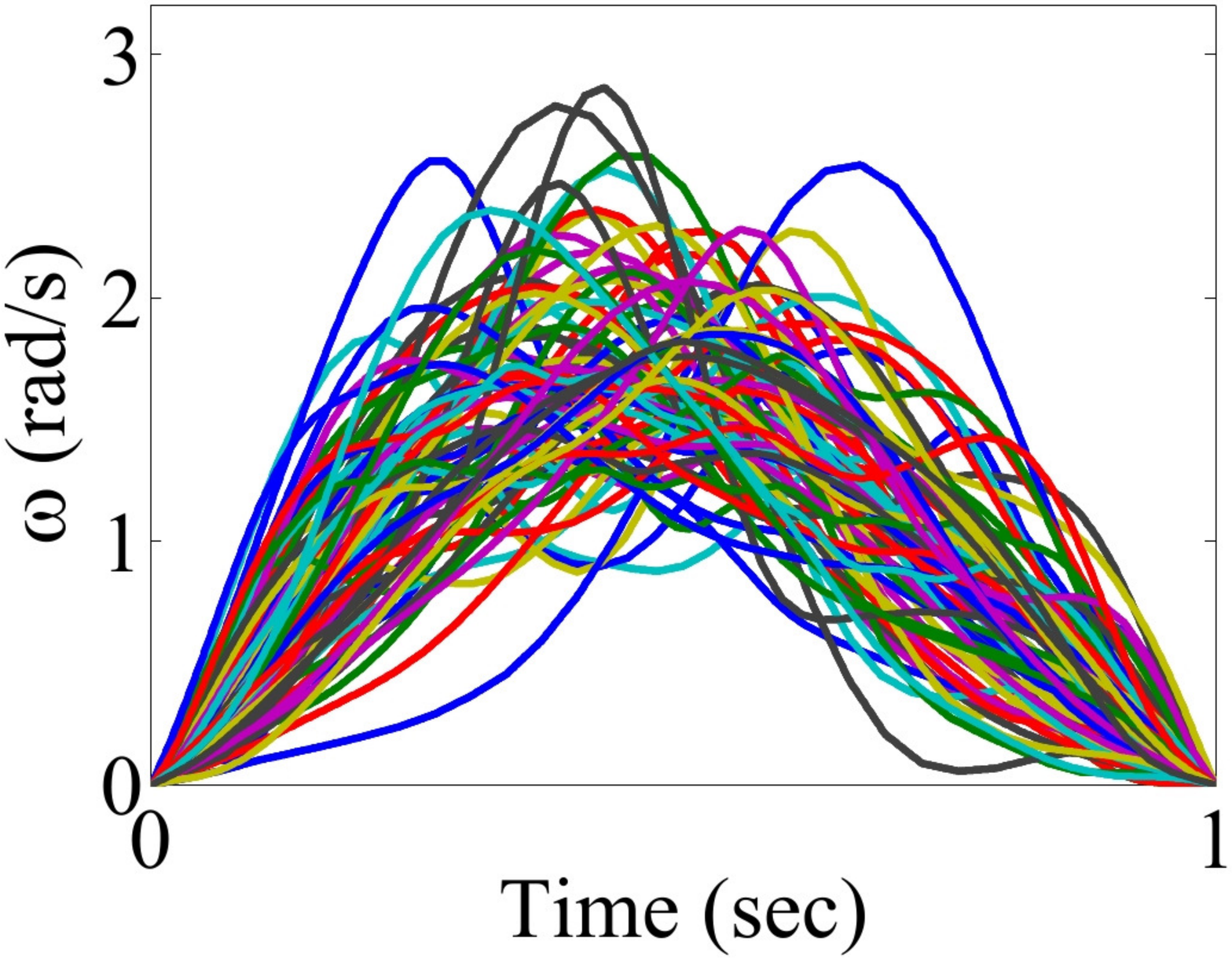}
	\hfill
	\includegraphics[width=.47\columnwidth]{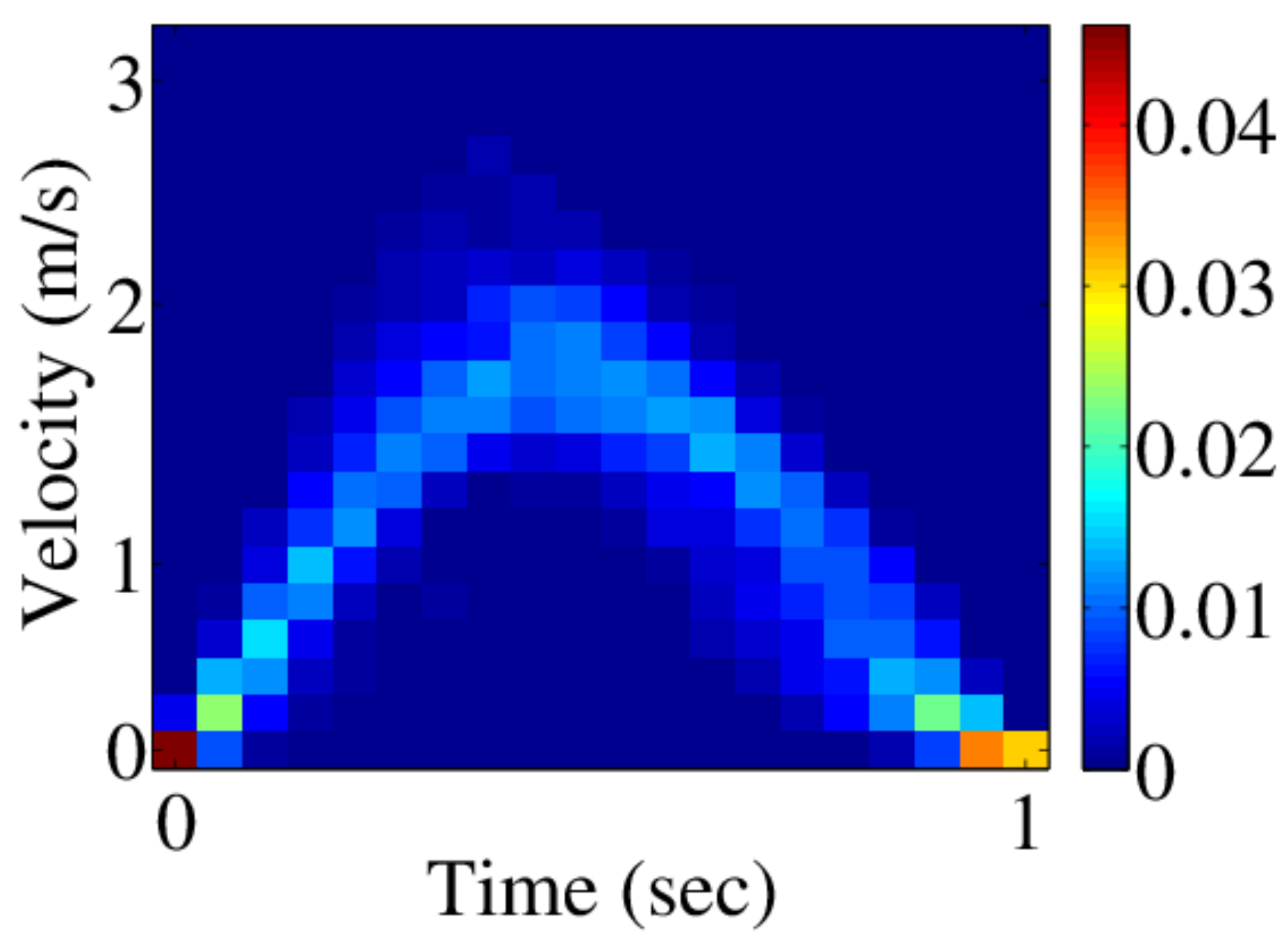}
  } \\
  \subfloat[\textit{
rotate} primitive \label{subfig-1:W}]{%
	\includegraphics[width=.47\columnwidth]{Signl-W-eps-converted-to.pdf}
	\hfill
	\includegraphics[width=.47\columnwidth]{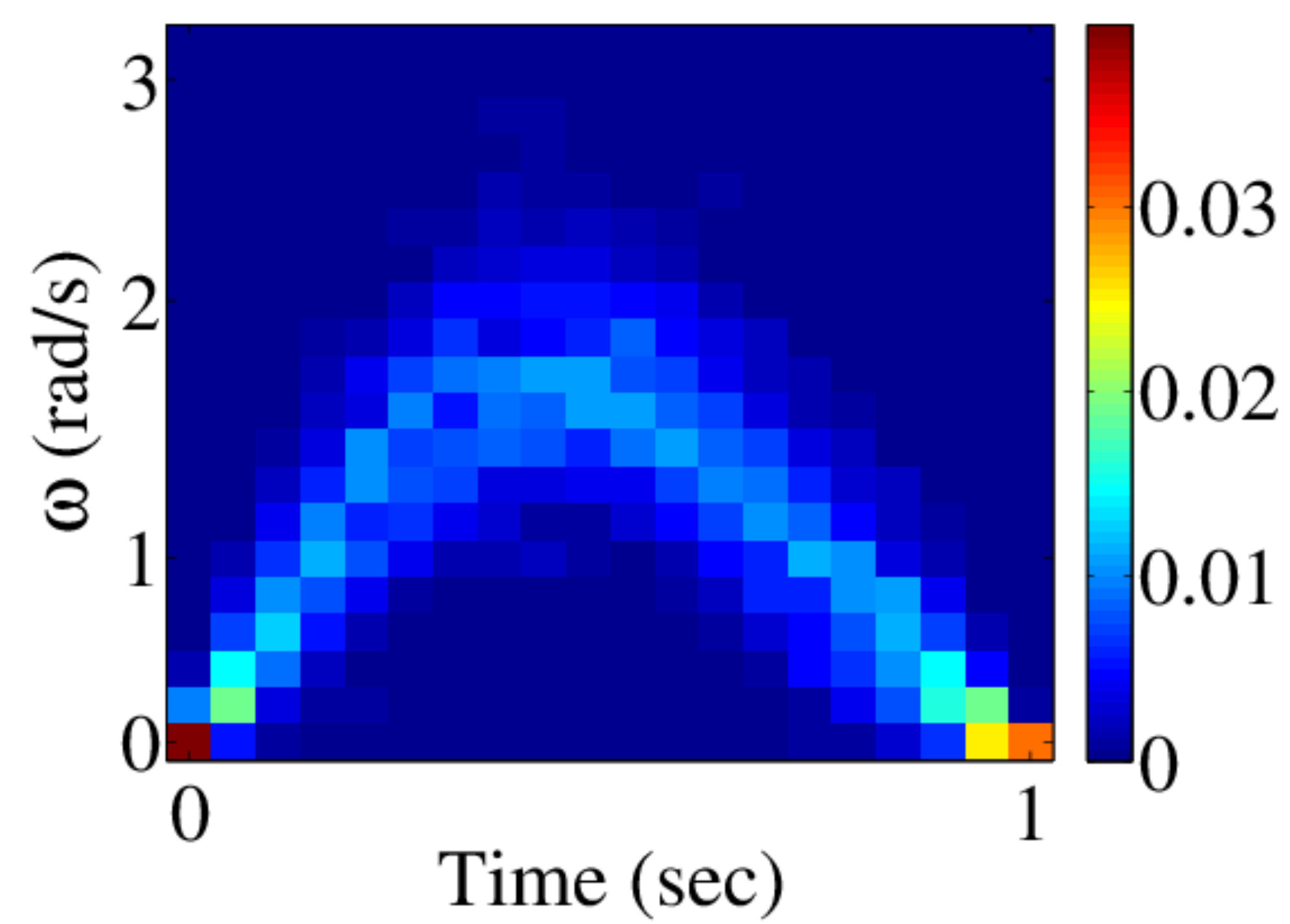}
  }
  \caption{Time trajectory and 2D histogram for the normalized manipulation primitives}
  \label{fig:PM-Results}
  \vspace{-6mm}
\end{figure}

To obtain the motion profile for~\textit{reach} and~\textit{rotate} primitives, we conducted
 \textit{MotionEXP} human study (see Fig. \ref{fig:expdirectories}) with eight healthy adult subjects. Each subject was asked to perform each primitive motions repeatedly wearing the data glove with an IMU mounted on the back of the glove (see Sec. \ref{Sect:valEXP} for more details). For the \textit{reach} primitive, the subjects were
instructed to move their hands in different directions and for various
distances (in rang of $30-50$ $cm$), at least ten times. The subjects performed the task either
empty-handed or while carrying an object. Various objects were
involved in the experiments including a mug, silverware, and a plate.  For the \textit{rotate}
primitive, the subjects were instructed to rotate their hands while
keeping their arms steady. The subjects were asked to rotate their
hand around the three principal axes of the hand (that also corresponded to
the axes of the IMU frame). Each subject performed at least ten
rotations around each axis (in rang of $0.7-3.14$ $rad$).

\begin{figure}[t]
\vspace{-6mm}
   \subfloat[\textit{reach} primitive \label{subfig-2:V}]{%
	\includegraphics[width=.47\columnwidth]{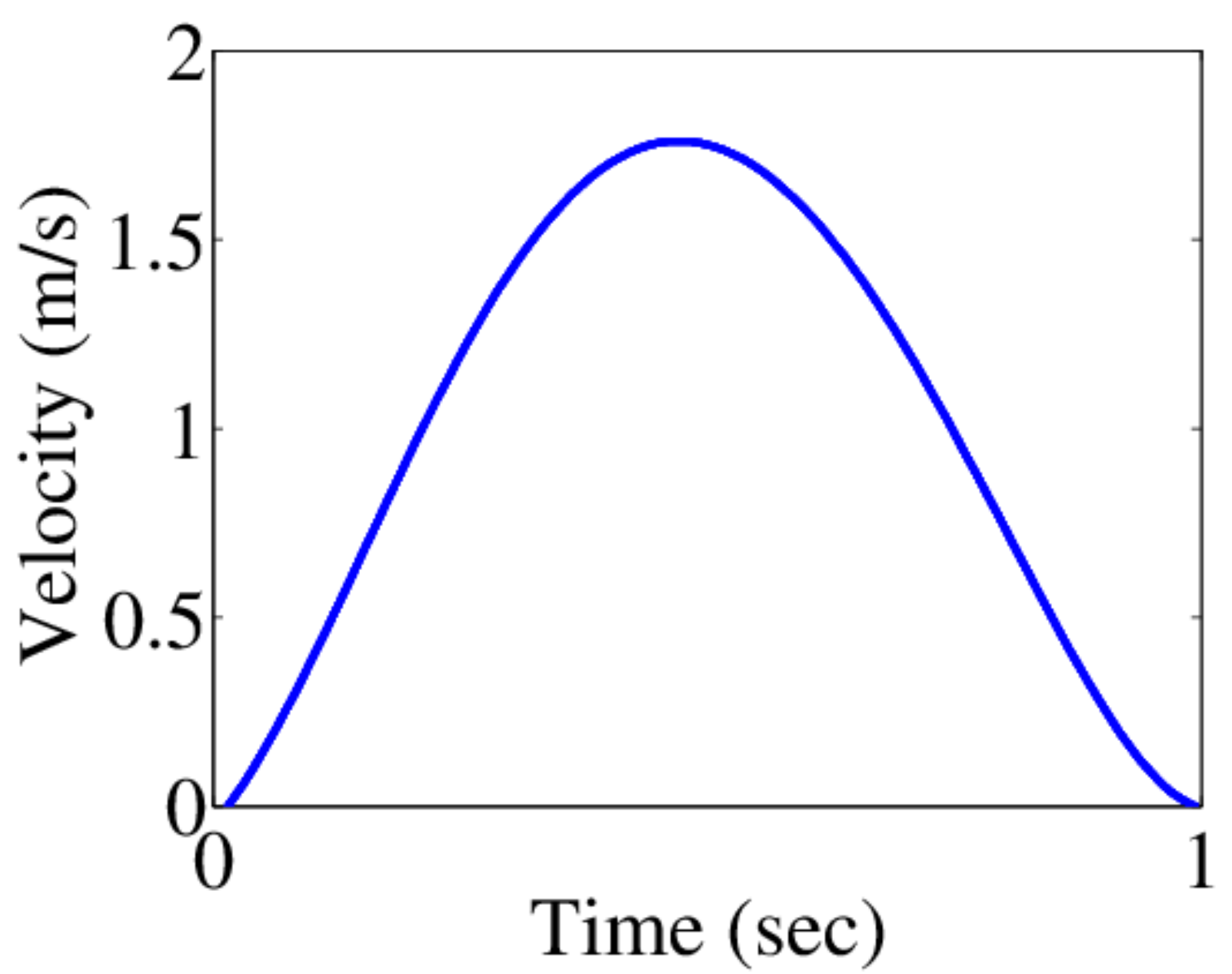}
  }
  \hfill
  \subfloat[\textit{rotate} primitive \label{subfig-2:W}]{%
	\includegraphics[width=.47\columnwidth]{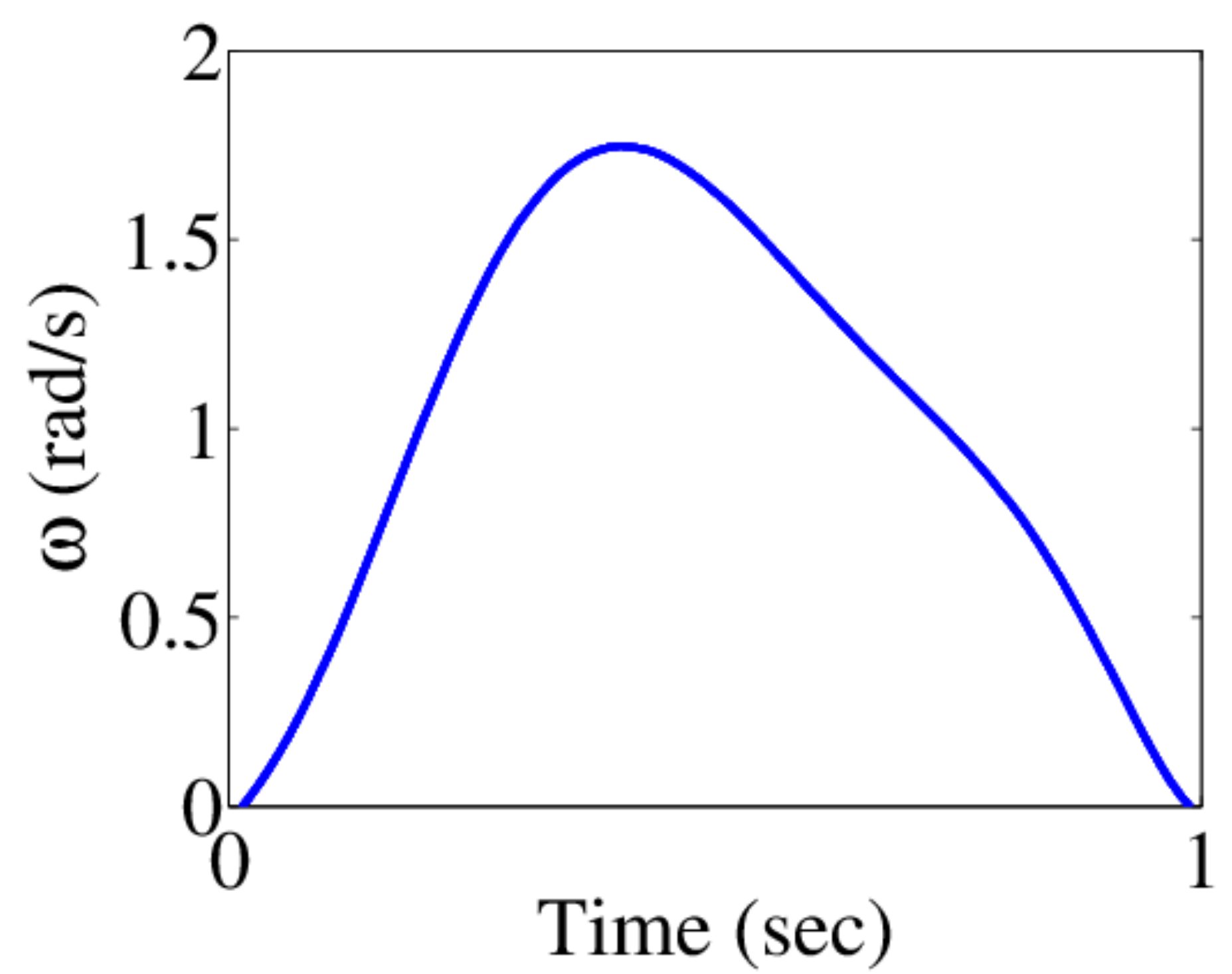}
  }
  \caption{Bell-shaped profiles for the \textit{reach} and \textit{rotate} primitives obtained from the human study}
  \label{fig:PM-Profiles}
\vspace{-4mm}
\end{figure}

After discarding the corrupted signals, we obtained
87 signals for the \textit{reach} primitive and 77 signals for the~\textit{
rotate} primitive. Fig. \ref{fig:PM-Results} shows the normalized velocity
trajectories in time domain for each primitive. To better illustrate the trajectory
profiles, the 2D histograms of these signals are also shown. As can be seen, the histograms roughly have a bell-shaped profile and they therefore in general agree with the predictions of the models in the literature~\cite{MinJerk_FlashHogans1985,MinTorqueChange_89,MinTorqueChange_99,MinVar_Wolpert}. However, to formally define the primitive features for the model for the \textit{reach} and \textit{
rotate} primitives, we decided to directly use the experiment data. Formally, we define the bell-shaped primitive profile as the most probable time trajectory in the collected data based on the 2D histograms in Figs. \ref{subfig-1:V} and \ref{subfig-1:W}. In calculating the
velocity profiles, we applied a smoothing filter on the maximum
probable trajectory. Fig. \ref{fig:PM-Profiles} shows the calculated
primitive profiles.\par

The bell-shapes defined in this way can be instantiated with
several parameters to match the physical trajectory of the hand.
These
parameters include $\omega$ which is the direction of the motion (or axis of
rotation), the magnitude of the motion (distance: $d$ or angle: $\theta$), start time ($t_{s}$)
and the duration of the motion ($T$) (see Tab. \ref{table:parameterslist}). Details on how the defined shapes are used for action recognition are provided in the next section. 

\begin{table}[b]
\centering
\vspace{-3mm}
\scalebox{1}{
\begin{tabular}[H]{|c|}
\hline
\textbf{Parameters of \textit{reach} and \textit{rotate} primitives}\\ \hline \hline
$\omega$ : axis of rotation/reaching\\ 
\rowcolor[gray]{.90}$\theta/d$: peak or angle of rotation/ distance of reaching\\ 
$T$: span of the primitive in time \\ 
 \rowcolor[gray]{.90}$t_{s}$: start time \\ \hline
\end{tabular}}
\vspace{0mm}
\caption{The list of parameters of \textit{reach}/\textit{rotate} primitives.}
\label{table:parameterslist}
\vspace{-2mm}
\end{table}

\begin{figure}[t]   
 \subfloat[Top View \label{fig:top}]{%
	\includegraphics[width=.47\columnwidth]{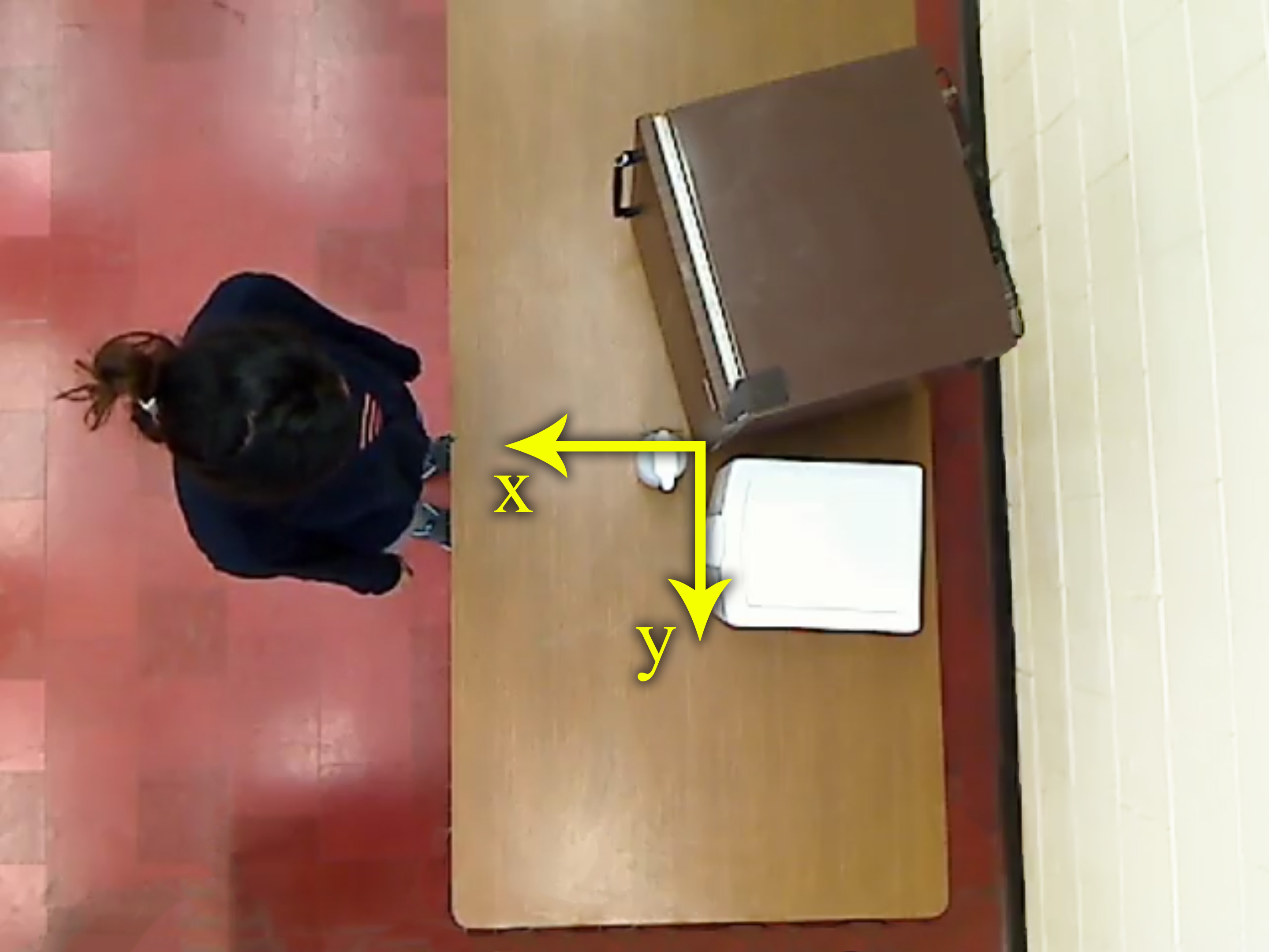}
  }
  \hfill
  \subfloat[Side View \label{fig:side}]{%
	\includegraphics[width=.47\columnwidth]{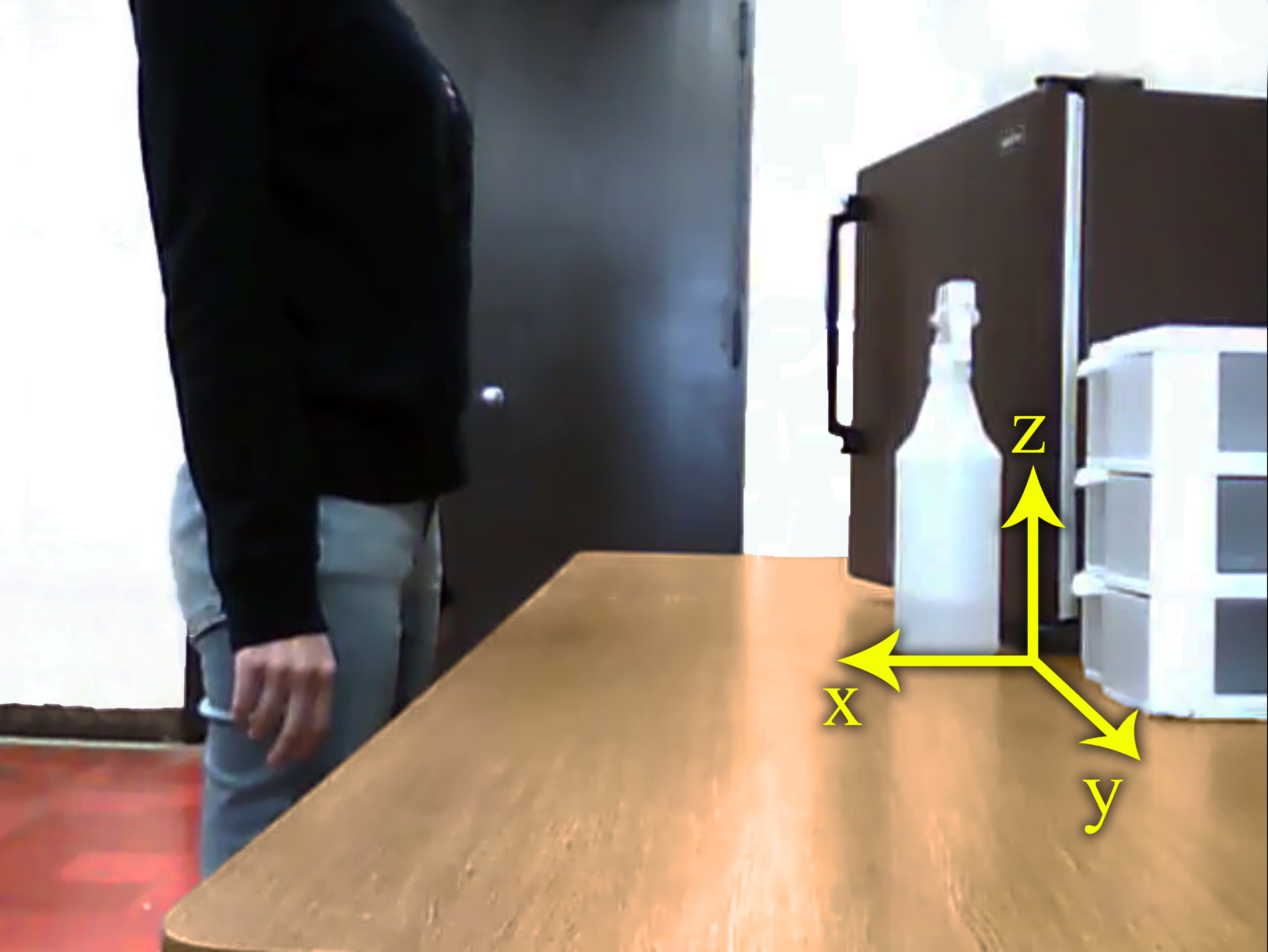}
  }
  \caption{The experimental setup, and the location of the coordinate frame on the table.}
  \label{fig:expsetup}
  \vspace{-2mm}
\end{figure}
\subsection{ValEXP Human Study: Validation of Primitives-based Method for Action Recognition}
\label{Sect:valEXP}
In this section, we present our method for identifying the manipulation primitives and their associated features directly without the need for annotation and classification.

Based on our analysis in section \ref{Sect:natexp}, we decided to use the following modalities for action recognition: hand linear velocity, hand rotational velocity, bending of the fingers, and pressure between the hand and the object. Next we describe how these signals are measured.

The experimental site consists of a table with the length of 185cm, width of 90cm, and height of 85cm. Two cameras were used, one placed directly above the middle of the table, and the other to the side of the table. The two cameras are used to track a color marker attached to the hand and are calibrated with $2cm$ resolution. Each camera provides a $2D$ motion trajectory of the hand with a sampling rate of $15$ frames/sec. The resulting trajectories were fused to extract the $3D$ trajectory of the hand in the reference frame that is set at the middle of the table, as shown in Fig.~\ref{fig:expsetup}. This trajectory was used to extract the linear velocity of the hand. Fig.~\ref{fig:expsetup} shows the top and side views of the experimental site.\par

\begin{figure}[t]
   \subfloat[Developed data glove \label{glove}]{%
	\includegraphics[width=.31\columnwidth]{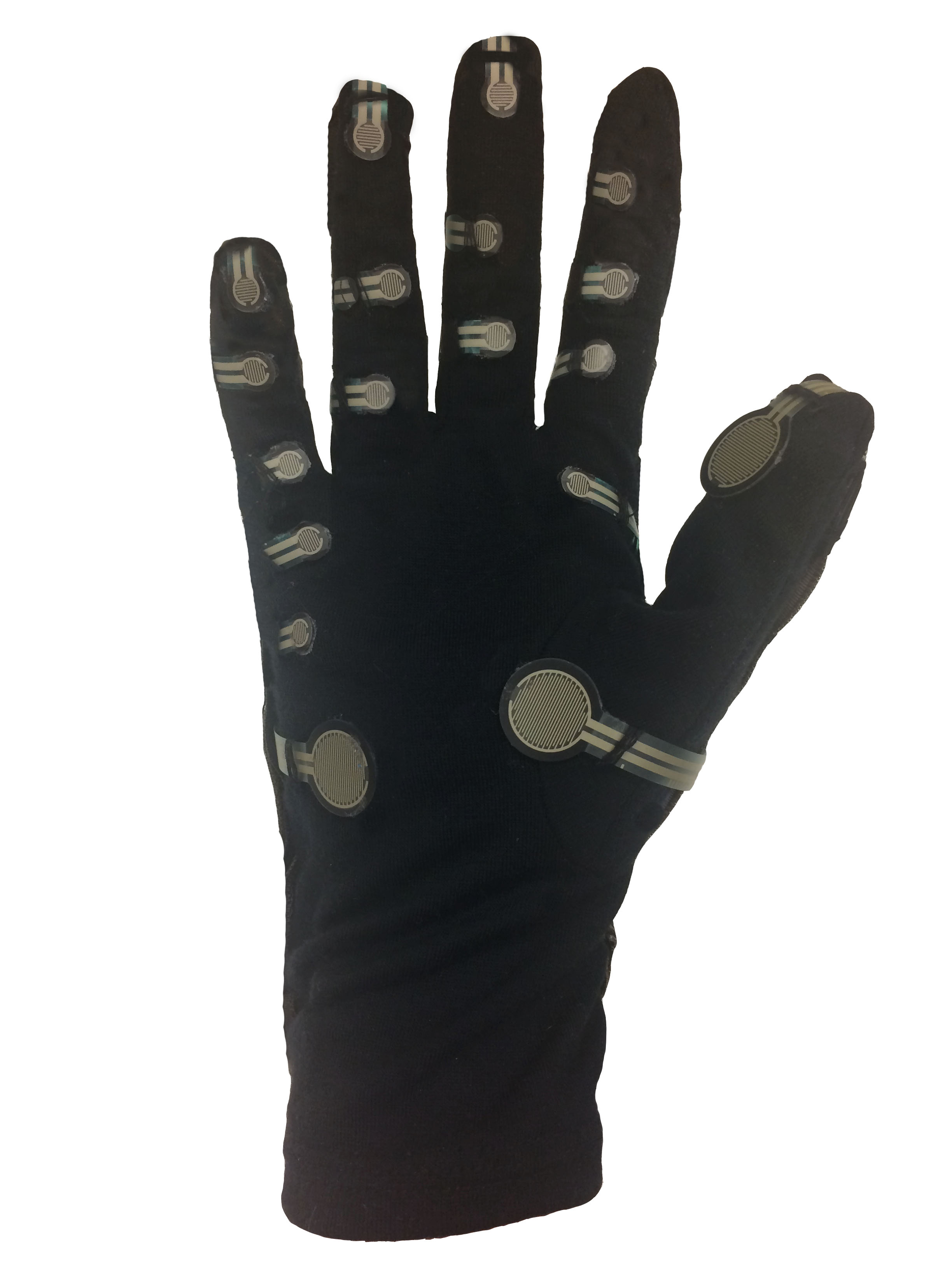}
  }
  \hfill
	\subfloat[FSR sensor locations \label{frontglove}]{%
	\includegraphics[width=.28\columnwidth]{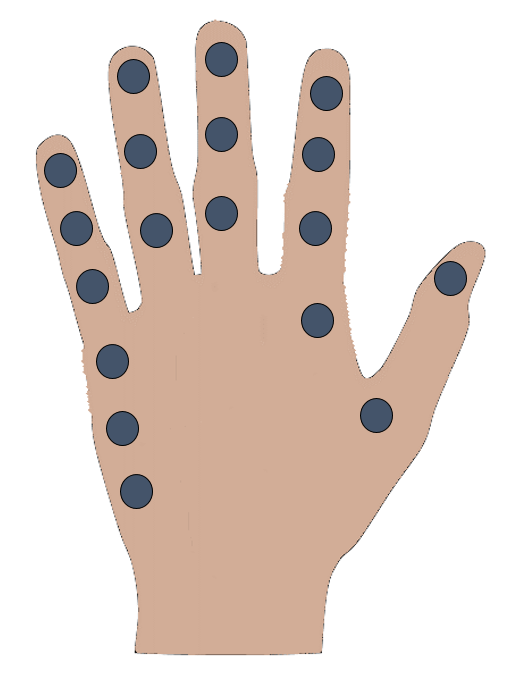}
  }
  \hfill
  \subfloat[Flex sensor locations \label{backglove}]{%
	\includegraphics[width=.28\columnwidth]{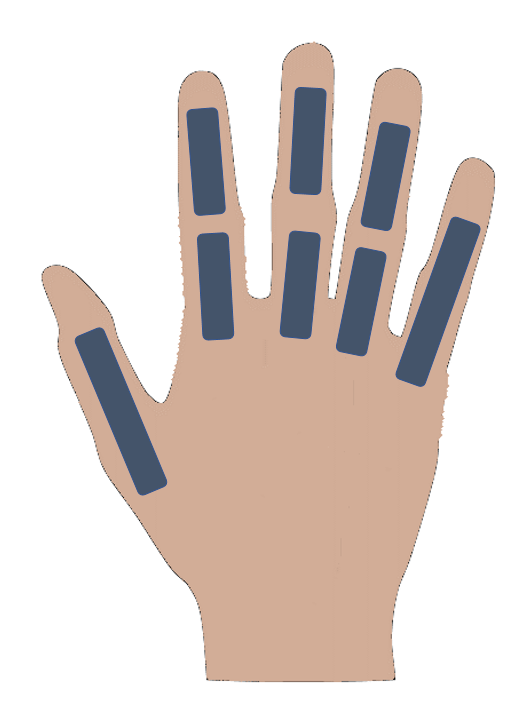}
  }
  \vspace{-2mm}
  \caption{The developed data glove equipped with pressure sensors, bend sensors, and an IMU.}
  \label{fig:dataglove}
  \vspace{-4mm}
\end{figure}

In order to measure other multi-modal signals, we developed our own data glove (see \cite{abbasi2019grasp}). The glove has eighteen flexible pressure sensors (FSR400-402 from Interlink Electronics Inc. \cite{FSR}) that cover all the active areas identified in~\cite{abbasi2016grasp} for different grasp types. 
Eight flexible bend sensors (Flexpoint~\cite{Flexion}) were placed on the back side of the glove to measure bending of the fingers. 
An Inertial Measurement Unit (Adafruit 10-DOF IMU) is attached to the back side of the glove to capture the angular velocity data. The data collection was conducted through an Arduino board (atmega 2560~\cite{Arduino}) with $100 Hz$ sampling rate. 
Fig. \ref{fig:dataglove} shows the data glove with the location of the FSR and Flex sensors.\par

\subsubsection{Primitive Features}
As discussed in Sec.~\ref{Sect:natexp}, primitives are too abstract to be directly use for action recognition. We thus define \emph{primitive features} that further characterize each primitive; these primitive features can in turn be used for action recognition. 

The primitive features are defined based on the collected data. We note two distinctive features of our approach. First, the features are multi-modal. In particular, we use pressure sensor data and flex sensor data in addition to motion data. Second, the features are defined in such a way that they can be identified in the data directly using signal processing, no learning is employed.

The modalities used for action recognition include four different types of signals: three linear velocities, three angular velocities, bending of the fingers, and the pressure between the hand and the object. The detailed description of all the components is given in Table~\ref{Table:Components}.

\begin{table}[b]
\centering
\vspace{-2mm}
\scalebox{1}{
\begin{tabular}[H]{|cc|}
\hline
\textbf{Multi-Modal Signal} & \textbf{Components} \\ \hline \hline
\multicolumn{1}{|c} {Hand linear velocity} & $v_x$, $v_y$, $v_z$\\ 
\rowcolor[gray]{.90}\multicolumn{1}{|c} {Hand angular velocity} & $\omega_x$, $\omega_y$, $\omega_z$ \\ 
\multicolumn{1}{|c} {Active areas force} & $F_1$,..., $F_{18}$ \\ 
\rowcolor[gray]{.90}\multicolumn{1}{|c} {Fingers joints bend} & $b_1$,..., $b_{8}$ \\ \hline
\end{tabular}}
\vspace{0mm}
\caption{All the collected signals during the experiments.}
\label{Table:Components}
\vspace{-4mm}
\end{table}

We next describe features of each modality in turn. We start with the \textit{grasp/release} primitive features. First, the norm of the $18$ dimensional vector formed by all the pressure senor values is computed. We will refer to this norm as the force signal. The average of the maximum of force signals in the train set is calculated. The features are then associated with the transition through a specific force level, and the direction of the transition. We quantized the force into three high, medium and low levels which are 75\%, 45\%, and 15\% of the calculated average. If the force signal crosses through one of the levels with positive slope, the primitive feature is recorded as \textit{grasp} primitive, otherwise if the force signal passes one of the levels while decreasing, it is \textit{release} primitive. The \textit{grasp} primitive features are shown by $p_{g_{l}}$, $p_{g_{m}}$, and $p_{g_{h}}$ and \textit{release} primitive features are $p_{r_{l}}$, $p_{r_{m}}$, $p_{r_{h}}$ which indicate the crossing level of low, medium and high respectively.\par

\begin{table}[t]

\centering
\scalebox{1}{
\begin{tabular}[H]{|cc|}
\hline
\textbf{Primitive} &\textbf{Primitive Features}   \\ \hline \hline
\multicolumn{1}{|c} {\textit{reach}} & $p_{v_{x^-}}$, $p_{v_{x^+}}$, $p_{v_{y^-}}$, $p_{v_{y^+}}$, $p_{v_{z^-}}$, $p_{v_{z^+}}$\\ 
\rowcolor[gray]{.90}\multicolumn{1}{|c} {\textit{rotate}} & $p_{{\omega}_{x^-}}$, $p_{{\omega}_{x^+}}$, $p_{\omega_{y^-}}$, $p_{\omega_{y^+}}$, $p_{\omega_{z^-}}$, $p_{\omega_{z^+}}$\\  
\multicolumn{1}{|c} {\textit{grasp}} & $p_{g_{l}}$, $p_{g_{m}}$, $p_{g_{h}}$\\ 
\rowcolor[gray]{.90}\multicolumn{1}{|c} {\textit{release}} & $p_{r_{l}}$, $p_{r_{m}}$, $p_{r_{h}}$\\ 
\multicolumn{1}{|c} {\textit{bend}} & $p_{b_{l}}$, $p_{b_{m}}$, $p_{b_{h}}$\\ 
\rowcolor[gray]{.90}\multicolumn{1}{|c} {\textit{extend}} & $p_{e_{l}}$, $p_{e_{m}}$, $p_{e_{h}}$\\  \hline
\end{tabular}}
\vspace{0mm}
\caption{The primitives features.}
\vspace{-4mm}
\label{table:features}
\end{table}

Similar approach is used to associate features with the finger bending signals. That is, first the norm of the vector of the sensor measurements is computed to obtain the composite bend signal. Next, the primitive features are defined by the composite bend signal and the direction of transition through the levels. As before, the high, medium and low levels are used. Transition through one of these levels in the positive (rising) direction corresponds to the \textit{bend} primitive features ($p_{b_{l}}$, $p_{b_{m}}$, $p_{b_{h}}$), while transition through one of these levels in the negative (falling) direction corresponds to the \textit{extend} primitive features ($p_{b_{l}}$, $p_{b_{m}}$, $p_{b_{h}}$) respectively.\par

To assign primitive features to the linear velocity signal, we use the bell-shaped motion profile for \textit{reach} primitive shown in Fig.~\ref{subfig-2:V} obtained from the \textit{MotionExp} human study. To extract the bell-shapes from the signal, first all the signal peaks with corresponding magnitude and location within the original signal are detected. Then, we assign the proposed profile for \textit{reach} to each peak at it's location with it's magnitude. In this way, the original signal is reconstructed by the resulted bell-shapes. By employing the optimization toolbox of MATLAB, the best approximation of the signals is obtained by finding the optimal duration of each bell-shape. The cost function of the optimization function is the mean square error($mse$), defined for original signal and it's approximated ones with the bell-shape profiles. In this way, the \textit{reach} primitive features as well as their associated parameters are identified within the signal in each of the three directions ($x$, $y$ and $z$). We follow the same approach for the angular velocity signal and the proposed profile for \textit{rotate} primitive shown in Fig.~\ref{subfig-2:W}. We discard the bell-shapes that represent the movement with distance/angle of less than $10cm$/$0.5rad$.\par

To remove the dependency of the features on the magnitude, we only consider the direction (positive or negative) of the velocity. That is, each identified bell-shape is abstracted with one of two different symbols. For the linear velocity signal in $z$ direction ($v_z$), we present the two features as ``$p_{v_{z^+}}$:move up'' and ``$p_{v_{z^-}}$: move down''. Similarly, we defined `$p_{v_{x^-}}$: move forward'' and ``$p_{v_{x^+}}$: move backward''. For side motion in $y$ axis, we have two possible primitive features $p_{v_{y^+}}$ and $p_{v_{y^-}}$.
For angular velocity signal, we consider two primitives features for each direction, one for negative and one for positive: $p_{{\omega}_{x^-}}$, $p_{{\omega}_{x^+}}$, $p_{\omega_{y^-}}$, $p_{\omega_{y^+}}$, $p_{\omega_{z^-}}$, $p_{\omega_{z^+}}$. The final list of manipulation primitives with their corresponding primitives features is provided at Table \ref{table:features}.

Further, we combine the bell-shapes in different directions which occur at the same time or very close to each other. To do so, given a bell-shape $p_i$ in a particular direction, if a bell shape $p_j$ in another direction occurs within the first half of its duration, we combine the features [$ p_i, p_j$]. Since each combination represents a motion, the order of  the primitive features does not matter in it. Therefore we reorder them according to their lexicographical order. This is done separately for the \textit{reach} and \textit{rotate} primitive features.\par

\begin{table*}[!t]
\vspace{2mm}
\centering
\scalebox{1}{
\begin{tabular}{ccccccc}
\toprule
{Manipulation} & \multicolumn{3}{c}{Primitive Feature} & \multicolumn{3}{c}{Raw Feature} \\
 
{Actions} & HMM & FC-LSTM & Conv-LSTM & HMM & FC-LSTM & Conv-LSTM\\
\midrule

Close Drawer (CD)   &  1.00 & 1.00   & 1.00 &  0.67 & 0.67 &0.53\\
\rowcolor[gray]{.90}Open Drawer (OD)  &  0.93 &  0.89  & 1.00  & 0.74 & 0.62 &0.29\\
Close Cabinet (CC)  &  0.84  &  0.90   & 1.00  & 0.89 & 0.50 &0.50\\
\rowcolor[gray]{.90}Open Cabinet (OC)  &  0.89  &  0.82   & 0.95 & 0.90 & 0.64 &0.56\\
Pick/Place (P-P)  &  0.78  &  0.71   & 0.70  & 0.62 & 0.67 &0.32\\
\rowcolor[gray]{.90}Spray (Sy)   &  1.00  &  1.00   & 1.00  & 0.84 &  0.95 &0.43\\
Stir (Sr)  &  0.93  &  0.86  & 0.77  & 0.88 & 0.55 &0.22\\
\rowcolor[gray]{.90}Pour (Pr)  &  0.71  &  0.88  & 0.86  & 0.71 & 0.35 &0.71\\\hline
Overall & 0.88 & 0.88 & 0.91 & 0.79 & 0.62 &0.44\\
\hline
\end{tabular}}
\vspace{0mm}
\caption{Comparison of recognition result for three different sequential model with primitives features and raw data feature in terms of F1-score for each action.}
\label{table:actionstable}
\vspace{-3mm}
\end{table*}

Using the proposed library of manipulation primitives and their corresponding features, human manipulation action can be recognized by performing the following two steps: first one needs
to mark the multi-modal human data including motion and force with the defined primitive features (a total of 24 symbols). Next, a stochastic temporal model needs to be trained to recognize the sequence of these abstract primitive features to recognize the manipulation actions. 
It should be added that these features remove the dependency of the recognition system to the absolute value of the continuous signal. \par

\subsubsection{Recognition Model}
\label{sec:primtive-models}

To validate the proposed methodology, we collected human multi-modal data for a subset of manipulation actions observed in ADLs. We considered a set of eight common object manipulation actions that had been observed in the \textit{NatExp} human study: {open/close a drawer, open/close a cabinet, pick and place a ladle, stir with a ladle, spray a sprayer bottle, pour a mug}. The first column in Table~\ref{table:actionstable} lists the selected manipulation actions. To simulate the environment in an apartment, we placed a drawer and a cabinet on the experiment site along with a mug, a sprayer bottle, and a ladle.\par

Five participants were recruited for the experiment. They were asked to perform each of the manipulation task for four to seven times in an arbitrary order. They performed the actions with an arbitrary starting and ending position, but the initial orientation was always fixed in order to be able to interpret the angular velocity provided by the IMU. For example, in the ``pick and place'' action, the subject chose the initial position, reached for the object, moved it to an arbitrary new position and retrieved his/her hand. All the actions started with reaching for the object and ended with retrieving the hand. We applied the primitives features extraction method, and found the sequence of them inside each trial. After removing the corrupted trials, overall $264$ trials were recorded during the experiment. The collected data for three subjects are used as training set, and the two remaining subjects as the testing set. We train three different sequential models HMM, FC-LSTM, and Conv-LSTM. The abstracted signals via the primitive-features are used as the input of the sequential models. The details of the training of ach model is provide in the following.\par

Hidden Markov Model (HMM) are among the most common classifiers for small feature-space sequential data. In this approach, each class, i.e. each manipulation action is modeled via a single HMM. Two different topologies (left-right (Bakis) and fully connected (ergodic)) with different number of states are experimented. The one with maximum likelihood is selected for each action. The predicted labels for testing set determined by the maximum likelihood among the HMM models. \par

The FC-LSTM and Conv-LSTM are both simple neural network based model including the Long short-term memory (LSTM) layer. In the former one, we employ Embedding (dimension: $500$) and Mask layer to pre-process the data, and one fully-connected (FC) layer ($300$ nodes) with ReLu activation followed by three LSTM layers (dimensions: $325$, $220$, $220$) with tanh and hard-sigmoid as activation and recurrent activation respectively and the final FC layer with $8$ nodes (each represents a manipulation action) and sigmoid activation for classification. In the latter network, we employ the Embedding layer along with convolutional layer (100 filters, kernel size: $3$), a max pooling layer, a LSTM layer (dimension:$100$), and the same output layer. For both of the networks, Dropout layer with rate of $0.5$ is used before LSTM layer and the loss function is categorical cross entropy and optimizer is Adam with learning rate of $0.001$. \par

The performance of the aforementioned trained models is presented in Tab. \ref{table:actionstable}. Additionally, to prove the effectiveness of using primitive-based model compared to using the raw measurements signals, we trained the same three models with raw measurements signals this time. Since the raw data signals include continuous values, we employed the HMMs with Gaussian Mixture Models (GMMs). Moreover, in LSTM-based models the Embedding and Mask layers (if existed) are replaced with a FC layer ($500$ nodes) to process the continuous data. In Conv-LSTM network, in addition to convolutional layer (dimension: 300), three LSTM layer (dimensions: 325, 220, and 220) are used. We used Dropout layer with ratio of 0.5 before each LSTM layer. The hyper-parameters of the models are optimized accordingly.

\begin{figure}[t]
\centering

    \includegraphics[width=0.9\columnwidth]{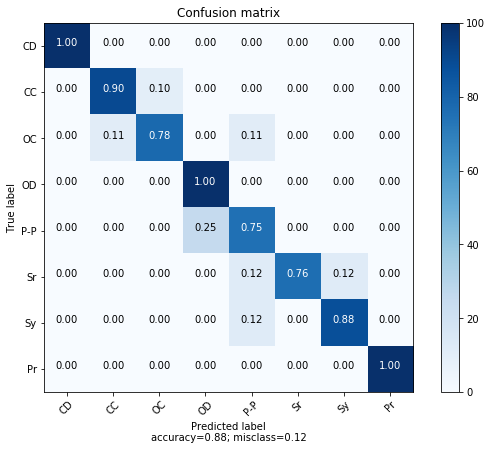}

    \caption{The confusion matrix of FC-LSTM trained with primitive-based model}
  \label{fig:confusion}
  \vspace{-6mm}
    \end{figure}

 The results clearly reflect the success of the proposed primitive-based method compared to using raw data. It also implies that using raw data without any treatment like extracting hand-crafted features is not sufficient for training the recognition models. \par

The confusion matrix of FC-LSTM model via primitive-based method is provided in Fig. \ref{fig:confusion}. The results from this model and others that trained using the same method imply that the primitive-based method abstract the raw continuous measurements data space into $24$ different meaningful and parameter-free symbols, and show better performance in recognition of human manipulation actions. Based on the confusion matrix, the ``pick and place a ladle'' actions are confused with ``pouring a mug'' and ``string a ladle'' actions because these actions are very close to each other in terms of generated signals. 
\section{visual-data-flow approach}
\label{Sect:data-driven}

In this section, we propose the purely data-driven method to recognize the human manipulation actions. We combine the extracted visual features from the videos with other modalities signals including motion of the hand (angular velocity and linear velocity), bend (fingers joints angles), and force (resulted pressure from active areas of the hand) to train the classifiers to perform the action recognition. \par

\begin{table*}[t]
\vspace{2mm}
\centering
\scalebox{0.9}{
\begin{tabular}{ccccccccc}
\toprule

{Classifiers} & \multicolumn{4}{c}{Resnet-v2-152} & \multicolumn{4}{c}{Inception-Resnet-v2} \\
 
{} & SVM & MLP  & FC-LSTM2 & Conv-LSTM2 & SVM & MLP  & FC-LSTM2 & Conv-LSTM2 \\
\midrule

20 frames & 83.36 & 81.78&81.66&89.24&82.51&82.54&89.69&84.53\\
70 frames & 85.97&84.26&82.23&91.06&87.42&87.06&89.13&89.39\\

Combined &86.29&88.71&88.50&90.65&87.42&93.92&91.17&92.57\\
\hline
\end{tabular}
}
\caption{Recognition result for different non-sequential and sequential classifiers trained with extracted visual features for 20 frames, 70 frames and combined features (with 70 frames) in terms of overall f1-score.}
\label{table:actionstable-vision}
\vspace{-4mm}
\end{table*}

\subsection{Visual Features from Deep Convolutional Neural Networks}

During the recent years, Deep Learning is used as a powerful technique for image classification tasks in computer vision; and various Deep Convolutional Neural Networks (DCNN) architectures have been proposed for it. In this paper, we take advantage of two recent proposed state of the art architectures for ImageNet Large Scale Visual Recognition Competition (ILSVRC) \cite{russakovsky2015imagenet}, and extract visual features from the pre-trained checkpoint. We employ \emph{Inception-ResNet-v2} which is the state of the art among the \emph{Inception} series networks. The difference between this network and previous Inception architectures is addition of residual connection to inception blocks which improves the learning~\cite{szegedy2017inception}. We also exploit a recent network from \emph{ResNet} networks, \emph{ResNet-v2-152} which has $152$ layers. The primary version of these deep residual networks are introduced by \cite{he2016deep} and proven to improve learning on image classification. In the second version series network, batch normalization has been added to weight layers~\cite{he2016identity}. All the networks are implemented by python under TensrorFlow platform.\par

To prepare the data for feature extraction from the pre-trained networks, we first transform all the videos to the frame images. We select the videos from side view camera. Afterward, the images for each trial are fed into each network separately. Since we are only interested in features extraction, the output of the last fully connected layer which does the classification is not useful, therefore, we use the fully connected layer before the last layer. The numbers of the nodes in this layer for \emph{Inception-ResNet-v2} is $1536$ which results in the same amount of features per image. Due to the reason that the layer before last layer in \emph{ResNet-v2-152} is not fully connected layer, we extract the features from the last pooling layer which result in $2048$ features per image.\par

After the feature extraction from each frame of the video of the trials, we utilize them to train the classifiers for human manipulation actions recognition. Two methods are employed for training the classifiers: non-sequential and sequential.\par

In order to train the non-sequential classifiers based on the extracted visual features, we need to select the particular number of frames from each recorded video in each trial, i.e. the number of features from all trials should be the same. To do that, we exploit an automatic clustering algorithm, K-means~\cite{hartigan1979algorithm}, to group the images into clusters, and the frames closest to the centroids are selected as the representatives of the video. Two different numbers ($20$ and $70$) are tried as the number of the clusters (K). In other words, we select $20$ and $70$ frames per trial. In non-sequential method, the extracted visual features for selected frames are concatenated for each video based on their sequence. Two different classifiers including Support Vector Machines (SVM) and a neural network model (Multi-layer Perceptron (MLP)) are trained for action recognition using the concatenated visual features. \par

In addition to non-sequential method, we train two sequential models, FC-LSTM2 and Conv-LSTM2 networks. The former network contains two LSTM layers with tanh and hard-sigmoid as activation and recurrent activation (dimensions: $500$, $220$) and a FC layer with ReLu activation ($400$ nodes) in between. In addition, two Dropout layers (ratio: $0.5$) are added after FC layer and the last LSTM layer. The same as the previous networks, the last FC layer which is responsible for classification has $8$ nodes and sigmoid activation function with the same training hyper-parameters.

The latter network contains one convolutional layer with 300 filters and kernel size of 3 followed by max pooling layer and two LSTM layers (dimension: $300$, $200$). The last layer is FC layer with $8$ nodes for classification. Two Dropout layers are added before each LSTM layer. The training parameters are the same as the training parameters mentioned in section \ref{sec:primtive-models}. \par

The result of the recognition for both feature sets extracted from the two networks with $20$ and $70$ frames is provided in the first two rows of Table \ref{table:actionstable-vision}. As the results imply the vision-based features are sufficient for action recognition purpose to some extend. This feature extraction method does not require training the deep models, therefore it does not need a huge dataset for training and is particularly good for small datasets. The performance of the models are close to those of the primitive-based models. In the sequential models for both feature sets from two different networks, the performance of the models with $20$ frames and $70$ frames are close to each other (except the Conv-LSTM2 with Inception-Resnet-v2 features), however in the non-sequential models the performance of the models with 20 frames is less than those using 70 frames. \par

\begin{table*}[!t]
\vspace{2mm}
\centering
\scalebox{1}{
\begin{tabular}{|c|cc|cc|}
\hline
&$f1score_{vision}$ &$f1score_{combined}$ &t-value &p-value \\
\hline

FC-LSTM2  &$84.20\pm(4.78)$ & $88.33\pm(4.81)$&-1.1148&0.3156 \\
\rowcolor[gray]{.90} Conv-LSTM2  &$89.56\pm(3.39)$ &$90.66\pm(2.50)$&-1.1284&0.3022\\
MLP  &$86.66\pm(5.33)$ &$91.31\pm(4.78)$&-2.3475&0.0657\\
\rowcolor[gray]{.90}SVM &$88.27\pm(4.47)$&$91.60\pm(4.21)$&-4.0304&0.0100\\
\hline
\end{tabular}
}
\vspace{0mm}
\caption{Significant test result on different classifiers using two different feature sets: vision-based features and combined features in terms of f1-score repeated for 6 times.}
\label{table:ttest}
\vspace{-3mm}
\end{table*}

\subsection{Multi-Modal data-flow Approach}

In previous section, we used only visual features for training the classifiers, in this section, we train the same models with the vision-based features along with the non-vision features including the hand linear and angular velocities, the norm of the force and bend signals (combined data). Here, we use the visual features extracted from videos for $70$ frames since they show similar or better performance than $20$ frames. The same as the previous section, for non-sequential method, we require to have equal number of features from each trials. Therefore, we concatenate the features from $70$ selected frames with the padded non-vision features to the maximum length of the dataset and train MLP and SVM classifiers with them. For sequential models, we used the same $70$ selected frames from videos, and added their corresponding non-vision features associated with that frame to them. We train the same LSTM-based models that explained in the previous section. \par

The train and the test set is similar to previous sections. The performance evaluation is presented in the last row of Table \ref{table:actionstable-vision}. The primary result suggests that adding the non-vision features may be beneficial for training some of the classifiers, however to generalize this finding, we need to show it statistically. The characteristics of different classifiers are distinct and they may process the features in completely different manner. In order to conclude the fact that adding non-vision features to the visual data improves the recognition performance significantly, we conducted significance test (t-test).\par

We used the visual features from ResNet-v2-152 with $70$ frames and trained all the aforementioned classifiers with three subjects and tested it with two remaining subjects. For each classifier, we replicated the procedure by shuffling the subjects in testing and training (cross-validation among the subjects trials). We trained each classifier in two modes: using combined features and using only visual features. The f1-score is utilized as the metric to measure the performance of the classifier in each mode. The performance results are provided in Tab. \ref{table:ttest}. \par  

The null hypothesis indicates that the performance of the recognition model using two different feature sets (visual features/combined features) does not significantly differ from each other. To implement the one sample t-test, we calculate the difference between the f1-scores resulted from each classifier in each mode. The hypothesized population mean is equal to zero. The resulted p-value and t-value are presented in Table \ref{table:ttest}. Based on the reported numbers for p-values, considering 0.05 significance level, the recognition rate improvement by adding the non-visual features is not statistically significant. The t-test results imply that the vision-based features are sufficient for three classifiers out of four. In other words, in the action recognition step, using the video recorded by cameras results in satisfactory performance. 

\section{Conclusion and Future Work}
\label{sect:conclusion}
In this paper, we studied how multi-modal data recorded during human object manipulation action can inform robot recognition of such actions, and how it can be used by the robot to learn how to reproduce them. We proposed two different approaches for action recognition, a primitive-based approach that is based on explicit physical models, and a visual-data-flow approach that only relies on the raw data.






In the primitive-based method, twenty four multi-modal primitives were defined using both physical models and the insights gleaned from the collected data. The multi-modal data used for this part of our study consists of hand motion data, applied forces as represented by the pressure patterns on the hand, and measurements of the bending of the fingers. The distinguishing characteristics of the proposed primitives is that they can be directly mapped onto the collected data, no classification needs to take place. In particular, the annotation and classifier training are not needed. Once the data had been segmented into the primitives, the sequence of the primitives was used to recognize the manipulation actions. Since the proposed primitives are explicitly modeled, they eliminate the over-fitting of the model and provide an abstract representation that does not depend on the absolute value of the signals. The comparison between the models trained with and without the primitives suggests that adding the information on the physics of the task to the recognition model is beneficial.\par

In the visual-data-flow method, we used vision in addition to the multi-modal data used previously. We extracted the raw visual features from recorded video using pre-trained deep convolutional neural network (DCNN), eliminating the need for developing a separate feature extraction methodology. The visual features were used to train several different classifiers both with and without using the multi-modal data. The results indicate that as concerns the recognition only, adding multi-modal data does not result in a statistically significant change in the classifier performance.\par

We show that both approaches produce a comparable classifier performance. This implies that image-based methods can successfully recognize human actions during human-robot collaboration. On the other hand, in order to provide training data for the robot so it can learn how to perform object manipulation actions, multi-modal data provides a better alternative.\par

Our results suggest several possible avenues for future research. While the visual-data-flow method can successfully recognize manipulation actions, in order to better control the interaction it may be necessary to also recognize the action primitives. We thus plan to study how the primitive-based method can provide the annotation labels to the visual-data-flow method and in turn generate a classifier without any human intervention. Another direction would be to move from object manipulation actions to higher-level tasks. This involves both choosing a task representation formalism as well as finding an appropriate planning and execution framework.
\bibliographystyle{ieeetr}
\bibliography{test.bbl}
\end{document}